\pdfoutput=1   
\documentclass{article}

\usepackage{arxiv}                     
\usepackage[numbers]{natbib}           

\usepackage[utf8]{inputenc}
\usepackage[T1]{fontenc}
\usepackage{hyperref}
\usepackage{url}
\usepackage{booktabs}
\usepackage{amsfonts}
\usepackage{nicefrac}
\usepackage{microtype}
\usepackage{xcolor}
\usepackage{tikz}
\usetikzlibrary{positioning,arrows.meta,calc}
\setcitestyle{square,numbers,comma}

\title{Govern the Model, Not Only the Data: Storage, Circulation, and Learning in Creative AI}

\author{%
  Phoenix Perry\\University of the Arts London\\\texttt{phoenix.perry@arts.ac.uk}
  \And
  George Simms\\University of the Arts London\\\texttt{g.simms@arts.ac.uk}
  \And
  Elizabeth Wilson\\University of the Arts London\\\texttt{e.j.wilson@arts.ac.uk}
  \AND
  Yasmine Boudiaf\\University of the Arts London\\\texttt{y.boudiaf@arts.ac.uk}
  \And
  Nick Bryan-Kinns\\University of the Arts London\\\texttt{n.bryankinns@arts.ac.uk}
  \And
  Tega Brain\\New York University\\\texttt{brain@nyu.edu}
  \AND
  R. Luke DuBois\\New York University\\\texttt{dubois@nyu.edu}
  \And
  Alix Rule\\New York University\\\texttt{aer342@nyu.edu}
  \And
  Rachel Meade Smith\\New York University\\\texttt{smithmeade@gmail.com}
  \AND
  Kelani Nichole\\New York University\\\texttt{kelaniatwork@gmail.com}
  \And
  Atharva Pravin Pawar\\New York University\\\texttt{app7633@nyu.edu}
  \And
  Rebecca Fiebrink\\University of the Arts London\\\texttt{r.fiebrink@arts.ac.uk}
}
\begin{document}
\setlength{\tabcolsep}{0pt}
\maketitle
\setlength{\tabcolsep}{6pt}

\begin{abstract}
Federated learning is increasingly presented as a privacy-preserving advance: personal data remain on the device, and only model updates are shared. It borrows the vocabulary of the federated social web, yet inverts its logic, distributing computation while the resulting model stays with whoever convened the training. We argue that federation is not in itself a remedy for extractive AI, because outcomes depend on who governs the data and the model and who has agency over the practices that shape them. We describe three layers at which a creative community can hold its work: storage, circulation, and learning. Examining artist-governed trusts, cooperatives, and consent infrastructures, we show that creator governance is established at storage and circulation but stops at learning: contributors can consent to training, yet have little say over the resulting model or its federation. We map the research space this opens, pairing technical open problems with the human questions from which they unfold. We propose four design principles for a creative data commons that governs models and their federation, not only datasets: govern the model, not only the corpus; make the terms legible at the moment of contribution; design for refusal as a first-class state; and decide stewardship in the open and account for it.

\end{abstract}

\section{Introduction: Two Federations}
Generative AI sits inside the everyday tools of creative practice, in text editors, drawing applications, and digital audio workstations through which artists, musicians, and writers make their work \cite{yafei2024generative}. Alongside large models trained on scraped web corpora, a yet to be negotiated architecture exists: federated learning, presented as a privacy-preserving advance because protected personal data never leave the device, and only the computed model updates are shared and federated \citep{mcmahanCommunicationEfficientLearningDeep2017}.

Federation is a protocol that can take on a number of forms; in this paper, we share two quite different architectures, which are routinely, if not intentionally, confused. The better-known FLOSS federated web3 technology lets communities run their own servers and interoperate as peers, through standards such as ActivityPub \citep{lemmerwebberActivityPub2018}. What is distributed is publication, and the community that runs a server decides what it hosts, who it federates with, and who it refuses \citep{gehlCaseAlternativeSocial2015,mansouxSevenThesesFediverse2020}. 

However, the federation of AI training embodies a second and distinct approach to federation. In this approach, devices compute model updates locally and send only those updates to a central server, which aggregates them into a single model \citep{mcmahanCommunicationEfficientLearningDeep2017,kairouzAdvancesOpenProblems2021}. What is distributed and federated here is computation, often involving computation on personal data. This architecture, in which local data remain on individual devices, while model aggregation and ownership are typically retained by a central coordinating organisation, in many ways inverses the logic of a FLOSS federated web. Google's Gboard, the longest-running deployment on the consumer scale, works exactly this way \citep{hardFederatedLearningMobile2018,xuFederatedLearningGboard2023}.
Similarly, Apple's co-option of differential privacy \cite{dwork2006differential} illustrates how concepts of privacy and federation have been absorbed into corporate AI infrastructure \cite{appleUnderstandingAggregateTrends2025}. Rather than enabling communities to govern their own data resources, such approaches optimise the extraction of aggregate insights. These real-world applications of federated learning present challenges, including high communication costs, statistical and system heterogeneity, persistent privacy vulnerabilities \cite{rossi2026federated}, as well as a lack of clarity of how on-device learning handles and maintains users' data \citep{cooray2025deep}. 

These two federation practices share an aim of decentralisation, but distribute different things: the first distributes data and offers governance over it; the second distributes model computation to extract personal data and retains ownership over the model. Using the same term for both makes their differences harder to make sense of. Consequently, federated learning often inherits the political standing of the federated web earned through community governance. The processing of the device is now both a market position and a technical method \citep{appleUnderstandingAggregateTrends2025}. The privacy claim is somewhat accurate in the scope of already individualising data rights such as GDPR \cite{maanengijsDataCommons2024}, since personal data stay only on the local device. However, data privacy is distinct from agency control over the practices that produce it. This broader question of agency has been raised in the creative sector, in evidence submitted to the UK copyright and AI consultation, and in sector assessments of revenue at risk\citep{CopyrightArtificialIntelligence,cisacEconomicImpactGenerativeAI2024}.

Our argument is that federation is a protocol, not a politics: whether it increases or
erodes a creative community's agency depends entirely on who governs the data and the
resulting model. Federated architectures are compatible with both building up governance
practices and extracting from them. For the case of this paper, we define creative communities as ``a group of people who come together around a shared challenge or theme to create, act, and share" \cite{ukriCreativeCommunities}. Our argument is informed by the practices of artists working with AI and data to propose some design principles that can build up the agency of creative communities and their capacities for collective governance. As such, our contributions are as follows.

\begin{enumerate}
\item Distinguishing the two main conflated federated paradigms, so that design decisions currently prescribed to people and communities can be made through deliberation (Section~\ref{sec:two}).
\item Differentiating \emph{storage}, \emph{circulation}, and \emph{learning} as governance layers for federated AI systems relevant to creative communities, where creator governance is established in the first two and currently absent in the third, drawn out through creator-governed federations such as the TRANSFER Data Trust and Serpentine's Choral Data Trust (Section~\ref{sec:gap}).
\item Mapping the research space, this opens, pairing technical frictions with artistic practice-based inquiries to formulate design principles for a creative data commons that can begin to govern models and their federation, not just datasets. (Sections~\ref{sec:agenda}--\ref{sec:principles}).
\end{enumerate}

\section{What creators already do}
\label{sec:two}

Artists often relate to machine learning in ways that deviate from and are counter to how AI systems are imagined and built. A recognisable practice has formed around small, self-curated datasets used to steer a model towards a particular aesthetic instead of general performance \citep{vigliensoniSmallDataMindset2022}. Anna Ridler photographed and hand-labelled ten thousand tulips for \emph{Myriad (Tulips)}, exhibiting the data set itself as work before training a GAN on it for \emph{the Mosaic Virus} \citep{ridlerMyriadTulips2018,ridlerMosaicVirus2018}. Jonathan Reus and the sound poet Jaap Blonk make the dataset live: in \emph{Bla Blavatar vs Jaap Blonk}, Blonk performs generated ``dataset poems'' on stage while his recorded voice is added to the training set to emerge a synthetic double\citep{reusBlaBlavatarBlonk2025}. Similar commitments run through the work of Sofia Crespo \cite{sofiacrespoNeuralZoo}, Eddie Wong \cite{wongUnknownPersonPostColonial2020, wongSpectresMay132022}, Magdalena Ty\.zlik-Carver \cite{tyzlik-carverCuratingFermentingData2022}, Jhave Johnston\cite{parrishReRitesHumanAI2019}, and Gabriel Vigliensoni \cite{vigliensoniRVAELiveLatent2022}, and through Holly Herndon and Mat Dryhurst's \emph{The Call}  \citep{herndonDryhurstTheCall2024}, which made a choral dataset assembled in collaboration with fifteen choirs\citep{ivanovaChoralDataTrust2025}. We do not present these as case studies, but draw from these practices to argue that a creative data commons can and already do operate at the artwork, gallery, and exhibition scale: artists curate, label, refuse, withdraw, and govern their data, which shapes outcomes both materially and conceptually. What remains underdeveloped are practices for governing and sharing the resulting AI models across creative communities.

\section{When the model is the material}
\label{sec:material}
A growing body of research \citep{caramiaux2022explorers} considers how the objects of machine learning practice including the models themselves can be understood as materials. Our point is not that a model is a material; it is what changes once a community, rather than a single maker, works it as one. 
Artistic practice makes especially visible a more general feature of human activity long recognised by philosophical pragmatists: people develop perception, judgement, and action capacities through sustained engagement with the materials and tools of their practice \cite{malafouris_how_2013}. In the context of the projects described above, the artists' data practices reveal how this dynamic operates. When models themselves become materials: interacting with models is not merely using a finished technical object; it is participation in a practice that forms both the system and the people working with it. Through data labelling, model training, and model sharing, artists actively reshape the creative and technical practices in which they are engaged, while also shaping the AI and data practices through which models are produced. By a community here, we mean a group that holds a shared body of work and could design, train, and govern a model together. 

Creative practice already treats trained models this way. Ridler describes a model surfacing patterns in her own drawing hand that she had not noticed \cite{ridlerGuestBlogPost2018}. Dadabots deliberately overfit short timescales and underfit long ones, so that a model trained on a single album generates within the grain of that album rather than escaping it \citep{vigliensoniSmallDataMindset2022}. Terrance Broad re-connects, bends and inverts the architectures and values of models \cite{broadNetworkBendingExpressive2021,broadAmplifyingUncanny2020} . In each case, the models are handled as material with grain, resistances, and capacity to be shaped, not as a determined pipeline optimised for accuracy or similarity. What these artists value is often what these systems background and remove and how they can negotiate these delineations through their situated practices.

We propose that when the model is cared for collectively, this makes room for its governance and matters to be shaped from many positions and desires. A community that curates a shared corpus together can make sense of how a model can be trained and designed, feeling how it responds, resists, prunes, re-weights, and refuses together. This offers a way of accessing and shaping AI and data's matters at the scale of collective practice and governance. This is a form of creative agency made known through artistic experimentation, but which is inaccessible today for predominantly infrastructural reasons: there is nowhere to manifest models on such collective and configurable terms. Governance of the learning layer is therefore not only a question of rights over training data, it orients how a model is intended to be shared, if at all, and under what conditions.

\section{Three layers, and where governance stops}
\label{sec:gap}

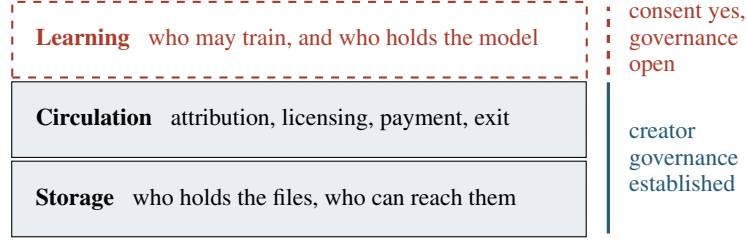
\begin{figure}[t]
\centering
\begin{tikzpicture}[font=\small, node distance=0pt]
\definecolor{held}{RGB}{38,90,120}
\definecolor{gap}{RGB}{170,60,45}
\definecolor{pale}{RGB}{233,238,242}

\node[draw, fill=pale, minimum width=7.6cm, minimum height=1.0cm, anchor=south west] (s) at (0,0) {};
\node[draw, fill=pale, minimum width=7.6cm, minimum height=1.0cm, anchor=south west] (c) at (0,1.05) {};
\node[draw, fill=white, minimum width=7.6cm, minimum height=1.0cm, anchor=south west, dashed, draw=gap, thick] (l) at (0,2.10) {};

\node[anchor=west] at ($(s.west)+(0.2,0)$) {\textbf{Storage} \; who holds the files, who can reach them};
\node[anchor=west] at ($(c.west)+(0.2,0)$) {\textbf{Circulation} \; attribution, licensing, payment, exit};
\node[anchor=west, text=gap] at ($(l.west)+(0.2,0)$) {\textbf{Learning} \; who may train, and who holds the model};

\draw[held, very thick] (7.9,0.05) -- (7.9,2.05);
\node[held, anchor=west, align=left] at (8.05,1.05) {creator\\governance\\established};
\draw[gap, very thick, dashed] (7.9,2.15) -- (7.9,3.05);
\node[gap, anchor=west, align=left] at (8.05,2.60) {consent yes,\\governance\\open};

\end{tikzpicture}
\caption{Three layers at which a creative community can hold its work. Creator-governed federations, such as the TRANSFER Data Trust and Serpentine's Choral Data Trust, have established governance at storage and circulation (Section~\ref{subsec:established}). At the learning layer, consent to training has been established in several cases, while governance of the resulting model remains under negotiation (Section~\ref{subsec:learning}).}
\label{fig:layers}
\end{figure}

We argue that the discourse of ownership and sovereignty within a federation misplaces the capacity of creative communities to access and shape their data and AI models from the bottom up. To help understand the possible interventions that creative communities can and have made here, we offer three layers in which data and models can be governed: storage, circulation, and learning (Figure~\ref{fig:layers}).

\textbf{Storage} is where the data or artefacts physically sit and who can access it. \textbf{Circulation} is the terms on which it moves: attribution, licencing, payment, and the right to withdraw. \textbf{Learning} is not only the technical act of training a model; it is the layer at which that act, and the model it produces, become open to collective governance: who may train a model, on which data and hardware and under what conditions, and who holds, adapts, and governs the model that results. The model and how it is trained can then enter the same loop of conversation and governance that storage and circulation already carry, rather than sitting outside it as a purely technical step. Existing approaches to governance within creative communities tend to focus on the first two layers of a system and not on the third. Importantly, within the learning layer, we distinguish consent to training from both the technical design and governance of the resulting model. Consent answers whether a contributor permits their work to enter a training process; model governance concerns what happens after that process: who may access, modify, adapt, deploy, redistribute, or withdraw the resulting model and under what conditions. For example, under GDPR a community could consent to training without having meaningful agency over the trained model or how it is subsequently used or modified, as long as any personal data is not identifiable from its interpolation.

\subsection{Storage and circulation: governance already established}
\label{subsec:established}
The TRANSFER Data Trust, a member-owned cooperative formed when a gallery dissolved in 2023 and returned its assets to its artists, federates storage rather than pooling it: each studio keeps its own archive, linked across peers for redundancy, so no contributor's work is absorbed into a single proprietary store \citep{nicoleHowCanData2025,transferDataTrust2025}. In the circulation layer, it redistributes a commission on sales annually and settles stewardship by quarterly vote of members. Stephanie Dinkins takes a critical approach to the storage layer in \emph{Data Trust}, encoding oral histories gathered with Black communities in San Jos\'e into soil and tree DNA, which makes durability and access material questions rather than contractual ones \citep{dinkinsDataTrust2025}. In cultural heritage, federated linked-data infrastructures have carried both layers for years \citep{rossenovaWikidataWikibaseComplementary2022}.

\subsection{Learning: consent established, model governance open}
\label{subsec:learning}
Serpentine's Choral Data Trust built a dataset with 15 UK choirs purpose-built for training, appointed a data steward, developed governance frameworks with legal advisors, and had external partners who are then able to train their models on it for Herndon and Dryhurst's exhibition \emph{The Call} \citep{ivanovaChoralDataTrust2025}. Fairly Trained assembles a consented public-domain corpus and trains a model only on it \citep{spawningSourcePlus}. Fairly Trained certifies models whose training data were licenced \citep{fairlyTrained}. Creative Commons' CC Signals proposes preference signals that can require a model trained on your work to be released openly \citep{creativecommonsCCSignals2025}. Moreover, traditional knowledge and biocultural labels developed by local contexts attach community authority and permissions as metadata that travel with the data and restrict its downstream use, prioritising collective rather than individual consent as defined by the CARE principles \citep{localcontextsLabels,carrollCAREPrinciples2020}.

While consent to training has been established, what remains under negotiated is governance of model and how they are trained. In the example of the Serpentine's Choral dataset, choirs consented to training; but do not decide who may adapt or adopt the model, on what terms it circulates, or what becomes of it once the exhibition closes. A certification audits a training set after the fact. A preference signal is a request that binds only those who choose to honour it, but little dialogue or resulting governance follows it. This is the open gap: consent to training has been established in several places, but governance of the trained model has not been well established. 

\section{Why this is a design problem and not only a technical one}

The absence of governance at the learning layer is not an oversight waiting for better aggregation. As Birhane argues \cite{birhaneAlgorithmicInjusticeRelational2021}, algorithmic harms are often framed as technical faults requiring technical remedies, while affected communities are excluded from making sense of their frictions or how they want to care for them. A relational understanding of personhood , data, and AI instead implies that  harm and repair are distributed across contexts and into emergent scales. For example, an artist can hold their files, be paid for their sales, but still have no agency over a model trained on their brushstrokes or practices, because each of those sits in a different system, governed by different terms, and exercised by different institutional actors.

Federating a data commons does not automatically erase power asymmetry, and in some ways it can accelerate it. Instead, we propose that the federation has the capacity to redistribute power. The capacity to run a node, steward a repository, read a licence, or maintain a model is unevenly distributed, and open source or shared data is open only to those with the education, time, compute, and social access to act on it \citep{liuOpenPanIdeologicalPanacea2023}. This is a labour question as much as a technical one: unpaid stewardship falls to whoever has capacity to give it, and the people an infrastructure is meant to sustain are frequently the people with the least capacity to maintain it. A commons designed without attention to existing power relations will reproduce inequalities in who can access and shape them.

\section{The research space this offers}
\label{sec:agenda}

Extending governance to the learning layer raises open problems.

\begin{itemize}
\item \textbf{Aggregation at small client counts.} Federated learning is currently engineered for thousands or millions of clients.  A creative community may consist of varying numbers of artists, studios, and galleries. What does aggregation look like when the client count is small, contributions are highly heterogeneous, and each contributor is identifiable and agential rather than anonymous in a crowd?  How do creators think about models in the first place, what imaginaries and which practices do they bring to a system they desire to contribute to?
\item \textbf{Model merging and provenance.} If a model is adapted from another model trained on several others, whose work does it rest on and how is that traced? What dialogues and practices around data are desired by creators for them to participate meaningfully in the decision making around that model, rather than nominally or passively?
\item \textbf{Withdrawal after training.} Contribution protocols are straightforward; withdrawal is not, since a model does not forget on request. What are contingent options and what could be promised? If withdrawal is a meaningful component of data governance, what would its equivalent at the model layer look like? What should an interface for data sharing, governance, and agency be like to be accessible, legible, and shaped by these communities?
\item \textbf{Feasibility without institutional computation}. Federated training is currently distributed across devices and orchestrated on a cloud scale. What is achievable at the scale of a small network and their computational capacities, and what does that rule in or out? How do the resulting approaches bridge a better understanding of creator's desires in technical, legal, and infrastructural design?
\end{itemize}

We offer these research directions not as discrete problems to be solved, but as areas where the technical, social, and practical frictions of federating AI learning must be worked through.

Federation also makes some things easier. Because contributions stay disaggregated, consent can be asked per contributor rather than per corpus, which is a granularity at which artists often already approach their work. Provenance becomes possibly tractable for the same reason: when a corpus contains data that documents who originated it at the level of the metadata, the question of whose work a model rests on remains answerable in principle, which is not possible for a model trained on a scraped web dump. Refusal begins to become a state a system can represent rather than an absence it cannot negotiate. However, for creative communities, the heterogeneity that makes small federations difficult to train is precisely the point. A centralised architecture treats the difference between twenty studios as variance to be averaged away and backgrounded, whereas a creative community treats these frictions as the generative reason to federate at all. Whether current aggregation methods can preserve that difference rather than smooth it is, to our knowledge, an open question. It cannot be settled from the technical side alone and will require close work with the creative communities whose practices define what that difference is worth preserving.

\section{Design principles}
\label{sec:principles}

We offer four principles to guide the design and governance of creative data commons. Our principles are situated because one that is useful to a systems builder is not necessarily relevant to a creator or regulator, each is addressed to specific constituencies.

\begin{enumerate}
\item \textbf{Govern the model, not only the corpus} \emph{(systems builders),} Contribution, attribution, adaptation, and withdrawal should be expressible for a trained model, not only for the data it trains on. TRANSFER's members can withdraw work from the archive; no equivalent operation exists for a model trained on it. CC Signals is an existing instrument, since it can ask that a model trained on your work be released openly \citep{creativecommonsCCSignals2025}, but it binds only those who choose to honour it. A system that governs only the data stops at the circulation layer.
\item \textbf{Make the terms legible at the moment of contribution} \emph{(systems builders, interface designers).} What a contributor is agreeing to should be acknowledged when they contribute, rendered in their own vocabulary. An artist adding to these systems should be able to shape these relations, forming their own terms of licence.
\item \textbf{Design for refusal as a first-class state} \emph{(system builders, creators);} Not contributing to, and acknowledging the backgrounds of practices, should be central to governance \citep{ciforFeministDataManifestNo2019}. Ridler exhibited her tulip dataset as an artwork; approaches that read it as a training input have already misclassified it. Indigenous data governance has precedent here, in labels that carry community permissions with the data rather than recording them elsewhere \citep{localcontextsLabels}.
\item \textbf{Decide stewardship in the open, and account for it} \emph{(communities, funders, policy),} Who maintains the commons, on what terms, and with what compensation should be an explicit, revisable decision. TRANSFER settles this with a quarterly member vote and a commission redistributed annually. Other collectives should consider multiple approaches, such as grant funding, grassroots organising, or a subscription-style model. Where it is left implicit, unaccounted stewardship falls to whoever can afford to give it, which reproduces the exclusions a commons is meant to escape. 
\end{enumerate}

For practitioners, the immediate implications are fairly relatable. An artist who trains a model on their own corpus today has few ways to share that model with peers on terms they set, and no vocabulary in which to state those terms. What is missing is not only the infrastructure through which such technical relations can be actioned, but also the vocabulary and legal instruments through which they can be expressed.

\section{Conclusion}

Federation is a site where creative agencies and practices can be deliberately extracted or embedded. The difference lies in whether communities can access the governance and shaping of their data and models and the collective agency they make sense of in action. These practices are creative and technical. Analysing storage, circulation, and learning shows that the creative sector has embedded approaches for two of the three layers, built through cooperatives and trusts organised by artists. The third area of learning remains to be negotiated. In future work, we aim to develop this layer through interviews, participatory co-design with UK and US creative communities, and a public database of open-source generative models documenting licences, training data, attribution, and governance frameworks. We are also prototyping and exploring community-governed approaches to model training and sharing.

Recent artistic practice makes the possibility of federation worth considering. The projects explored here open broader questions about how people whose data, judgments, and practices shape AI systems might claim collective agency over what those systems do and become. This paper raises the question of what it would mean for communities to govern and federate the learning process and resulting models. The challenges that surface are both technical and social: making provenance, refusal, modification, withdrawal, and stewardship actionable in learning making while acknowledging the situated practices and unequal capacities through which such governance operates.

\section*{Acknowledgments}
This work is part of \emph{Federated Data Commons for Creative Communities}, funded by the AHRC BRAID programme (grant number UKRI3355). We thank the artists and communities whose practice this paper cites for beginning the process of imagining  and actioning a future where AI and creative practice can co-exist equitably.

\bibliographystyle{unsrtnat}
\bibliography{references,references-additions-v3,references-missing}

\end{document}